\def\year{2017}\relax
\documentclass[letterpaper]{article}
\usepackage{aaai17}
\usepackage{times}
\usepackage{helvet}
\usepackage{courier}

\usepackage{url}
\usepackage{latexsym}
\usepackage{algorithm}
\usepackage{amsmath}
\usepackage{algpseudocode}
\usepackage{booktabs}
\usepackage{multirow}
\usepackage[dvips]{graphicx}

\begin{document}
%
\title{Improving Twitter Sentiment Classification via Multi-Level Sentiment-Enriched Word Embeddings}
\author{Shufeng Xiong\\
Pingdingshan University\\
Pingdingshan, Henan, China\\
}
\maketitle
\begin{abstract}
Most of existing work learn sentiment-specific word representation for improving Twitter sentiment classification, which encoded both n-gram and distant supervised tweet sentiment information in learning process. They assume all words within a tweet have the same sentiment polarity as the whole tweet, which ignores the word its own sentiment polarity. To address this problem, we propose to learn sentiment-specific word embedding by exploiting both lexicon resource and distant supervised information. We develop a multi-level sentiment-enriched word embedding learning method, which uses parallel asymmetric neural network to model n-gram, word level sentiment and tweet level sentiment in learning process. Experiments on standard benchmarks show our approach outperforms state-of-the-art methods. 
\end{abstract}

\section{Introduction}

Twitter, which is one of the biggest micro-blog site on the internet, has emerged as an important source for online opinions and sentiment indexes.
As a result of its massive, diverse, and rising user base, the containing opinion information in Twitter has been successfully used to many tasks, such as stock market movement prediction \cite{SiMukherjee-2023}, political monitoring \cite{PlaHurtado-2031} and inferring public mood about social events \cite{BollenMao-2024}. Therefore, excellent sentiment classification performance, the ability to identify positive, negative, and neutral opinion, is fundamental.

Researchers have proposed many approaches to improve Twitter classification performance \cite{DavidovTsur-2028,BarbosaFeng-2033,MukherjeeBhattacharyya-2030}.
Especially, recent advance in deep neural network demonstrate the importance of representation learning of text, e.g., word level and document(sentence) level, for natural language processing tasks \cite{LeMikolov-1632,CollobertWeston-1507,TangQin-1755}. Traditional  word embedding methods \cite{CollobertWeston-1507,MikolovSutskever-1335} model the syntactic context information of words. Based on them, \cite{TangWei-1500} proposed Sentiment-Specific Word Embedding (SSWE) learning method for Twitter sentiment classification, which aims to tackle the problem that two word with opposite polarity and similar syntactic role for sentiment classification task.  Following SSWE, \cite{RenZhang-2019} further proposed Topic and Sentiment-enriched Word Embedding (TSWE) model to learn topic-enriched word embedding, which considered the polysemous phenomenon of sentiment-baring words.

However, existing work only exploit Twitter overall sentiment label for learning sentiment-specific word embedding, although there are many state-of-the-art sentiment lexicons \cite{HuLiu-2036,WilsonWiebe-2035}, which list common sentiment-baring words with its polarity.
Existing work learn sentiment-specific word embedding by using distant supervised tweet polarity label based on traditional learning model.
Actually, traditional methods learn word embedding based on a local context model.
But, Twitter sentiment label belongs to global document level.
For exploiting tweet sentiment label, \cite{Tang-1658} and \cite{RenZhang-2019} assume that each word in an opinioned context window indicates the sentiment polarity of the context, and local context has the same polarity as the global sentiment of tweet.
In other words, they assigned the tweet level global polarity to its local context window without any adjustment.
On the other hand, word sentiment polarity from lexicon is still another useful information for sentiment classification \cite{PangLee-1837,Feldman-2037,SocherPerelygin-1940}.
Accordingly, a uniform framework to exploit multi-level sentiment label for learning word embedding is necessary.
How to effectively exploit both word and tweet sentiment label for learning word embedding, is still a challenge problem.

\begin{figure*}[!htb]
\centering
\includegraphics[width=0.9\textwidth]{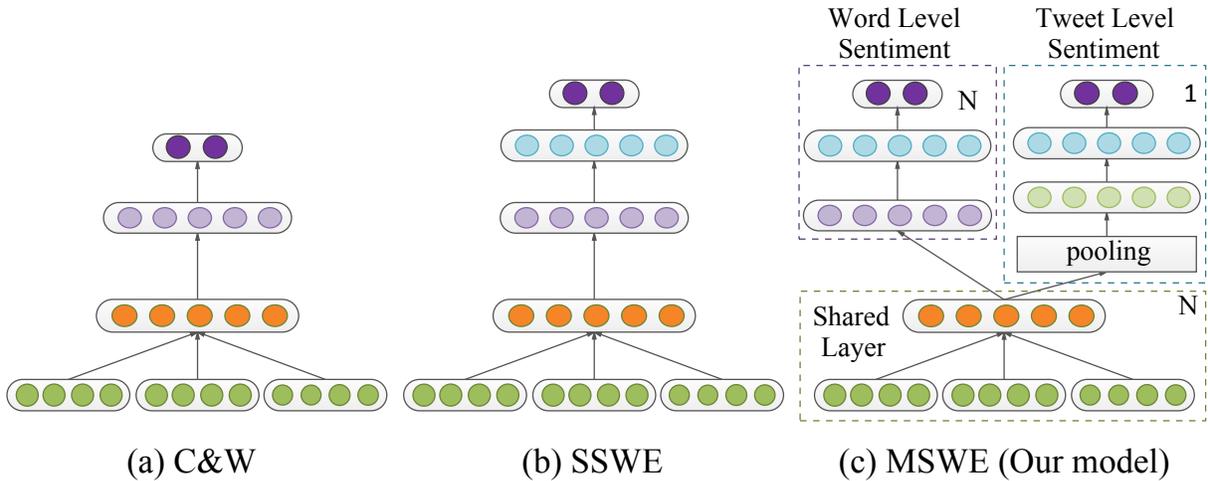}
\caption{Architecture of different word embedding learning methods.}
\label{fig:lambda1}
\end{figure*}

In fact, SSWE model combines multiple objectives, which are syntactic and sentiment learning, into one function. Moreover, TSWE model further encode topic, sentiment and syntactic into optimizing objective of neural network. However, most of the time, multiple objectives can not be directly optimized by a unified framework, i.e. word sentiment and tweet sentiment. Inspired by recent work for Multi-task deep Learning \cite{CollobertWeston-2013}, we propose to learn word embedding by exploiting multi-level sentiment representation objective optimization approach. Although multi-level sentiment representation can be seen as multiple tasks, they actually have two different inputs 1) word with its context and 2) the whole tweet, which correspond to word sentiment and tweet sentiment, while multi-task deep learning commonly has the same input.
Therefore, it is unable to directly use multi-task deep learning framework on this issue. 

For tackle this problem, we develop a Multi-level Sentiment-enriched Word Embedding (MSWE) model to learn word representation. MSWE consists of two two-parts asymmetric sub networks, which share a common linear layer and word representation layer. The two-parts sub networks are 1) several Multi-Layer Perceptron (MLP) networks and 2) one Convolution Neural Network (CNN), which are used to model n-gram and sentiment information of word level and sentiment information of tweet level, respectively. Our model is under the assumption that each word vector encodes its own sentiment and word composition encodes tweet sentiment, while SSWE/TSWE assumes each word vector indicate the tweet overall sentiment. Specifically, we feed words (with its context window) into MLP to model word level sentiment, which encodes the word its own sentiment. At the same time, the whole tweet is fed into CNN to model tweet level sentiment.

The contributions of this paper can be summarized as follows.
\begin{itemize}
\item We propose to encode both word level and tweet level sentiment information when learning sentiment-specific word embedding, which makes full use of existing sentiment lexicons and distant supervised Twitter corpus.
\item To address the multi-level sentiment representation objective optimization, we develop a novel word embedding learning framework, which employs two asymmetric sub networks to model two level sentiment information, respectively.  
\item We conduct experiments on standard Twitter sentiment classification benchmarks. The experiments results demonstrate that our method outperforms previous state-of-the-art approaches.
\end{itemize}

\section{Multi-Level Sentiment-Enriched Word Embedding for Twitter Sentiment Classification}

In this paper, we argue that it is important to jointly model both word level and tweet level sentiment information to learn good sentiment-specific word embedding. In this way, it not only made full use of existing sentiment lexicon resource, but also distant supervised Twitter corpus in a unified representation framework. In other words, n-gram and multi-level sentiment information are both encoded in word embedding, which can improve the sentiment classification performance.

Figure 1(c) describes the architecture of the representation learning network MSWE, which are used for encoding sentiment information in word embedding. The network takes tweet as input. First, MSWE model splits input tweet into several windows, i.e. $n$ windows. Then, $n$ window is the actual input of $n$ left sub networks and all the windows are the actual input of one right sub network. The left sub network outputs word level sentiment and n-gram information, and the right sub network output tweet level sentiment sentiment. We proceed to describe the network in detail.

\subsection{Sentiment-Specific Word Embedding}

Our model aims to learn sentiment-specific word embedding, which is more beneficial for twitter sentiment classification than common used word embedding.
Sentiment-specific word embedding model stems from C\&W model \cite{CollobertWeston-1507}, which learns word representation from n-gram contexts by using negative sampling. When learning the representation of a word $w$, they chose its contextual words $c$ in a window $t$ as positive sample, in which $w$ is in the center position of $c$. For getting negative sample, they alternate the word $w$ with a different word $\tilde{w}$ to form mutational context $\tilde{c}$. Its training objective is that the context $c$ is expected to obtain a higher language model score than the mutational context $\tilde{c}$ by a margin of 1. The optimization function is
\begin{equation}
\label{eq:g0}
loss_{ngm}(c) = \max (0, 1 - f^{ngm}(c) + f^{ngm}(\tilde{c})),
\end{equation}
where $f^{ngm}(\cdot)$ is the output of C\&W model, which represents the n-gram score of a context through out neural network.
The C\&W model architecture is shown in Figure 1(a). The bottom layer is lookup layer for represent $L$  $d$-dimension word vectors in vocabulary. All the words in context window are concatenated to form $[L_1,...,L_t]$, which is fed into a hidden linear layer. A $hTanh$ activation layer is used on top of the linear layer, its output is
\begin{equation}
a = hTanh(W_1*[L_1,...,L_t] + b_1),
\end{equation}
where $W_1 \in R^{h \times (t*d)}$ and $b_1 \in R^h$ is the model parameter, $h$ is the length hidden unit.
The n-gram score is computed by a linear transformation applied in $a$, 
\begin{equation}
f^{ngm}(c) = W_2*a,
\end{equation}
where  $W_1 \in R^{1 \times h}$ is the model parameter.

Although C\&W model is successfully applied in many NLP task, it is not effective enough for sentiment classification.
To address this problem, \cite{TangWei-1500} proposed a sentiment-specific word embedding SSWE model base on C\&W model.
They added tweet sentiment information loss into the objective function, which demonstrates the effectiveness of encoding sentiment information into word embedding for Twitter sentiment classification. The network structure is shown in Figure 1(b). There are some other variations, for example, TSWE (Topic-Enriched Word Embedding) encodes topic distribution into loss function of model and TEWE (Topic and Sentiment-Enriched Word Embedding) considers both topic and sentiment information \cite{RenZhang-2019}. Overall, all of them use multiple optimization objectives in their loss function.

\subsection{Multi-Level Sentiment-Enriched Word Embedding}
\label{sect:mswe}

Existing work demonstrate that sentiment lexicon is an important resource for sentiment classification \cite{PangLee-1837,Feldman-2037,SocherPerelygin-1940}. In particular, \cite{SocherPerelygin-1940} developed \textit{Sentiment Treebank}, which is based on word sentiment and further annotates the sentiment label of syntactically plausible phrase of sentences. By using this corpus, they train a RNTN (Recursive Neural Tensor Network) model for sentiment classification. However, the labelled phrase of \textit{Sentiment Treebank} is still limited, which can not contain all the available word combinations. Nevertheless, comparing with word combination, the basic sentiment-bared words are changed relatively less over time. Therefore, we propose to consider word level sentiment from lexicon for encoding the individual sentiment information, and simultaneously model tweet level sentiment for considering the composition sentiment information of words. 

An intuitively approach is adding a new optimization objective in the loss function as practice in SSWE and TEWE. But, there are $n$ word sentiment optimization objectives corresponding to $n$ input words in tweet and only one optimization objective for tweet level sentiment. Existing structure can not optimize these multi-level sentiment scoring objective. As shown in Figure 1(c), we exploit two sub networks for modelling two level optimization objectives by using similar framework as multi-task learning \cite{ChuOuyang-2016,CollobertWeston-2013,DongWu-2015}.

\textbf{Shared Units}\hspace{5pt} Two sub networks have shared units, each unit contains one embedding layer and one linear layer. Assuming there is an input tweet $D$ consists of $n$ context windows, embedding dimension is $d$ and hidden layer length is $h$. Then the shared unit number is $n$, each unit connects one left sub network, and the whole $n$ units connect to right sub network after a pooling operation. For each shared unit, the input of embedding layer is $t$ words in a window and the output is represented as:
\begin{equation}
x_{1:t} = x_1 \oplus x_2 \oplus ... \oplus x_t ,
\end{equation}
where $x_i,x_{i+1},...,x_{i+t}$ is the $t$ word in window $i$. A linear transformation is applied to $x_{i:i+t}$ to produce a new feature
\begin{equation}
e_i = f(W^1_1 * x_{i:i+t} + b^1_1),
\end{equation}
where $W^1_1 \in R^{(t*d) \times h}$ and $b^1_1 \in R^h$ are the parameter of linear layer.

\textbf{Word level-Specific Layers}\hspace{5pt} For left sub network (modelling word level sentiment and n-gram), a following $tanh$ activation layer outputs $a_1=hTanh(e_i)$ and two linear transformations output n-gram and word level sentiment predicted scores
\begin{equation}
f^{ngm} = W^1_2 * a_1,
\end{equation}
and 
\begin{equation}
f^{ws} = W^1_3 * a_1
\end{equation}
When training the model, we input window $c$ and its mutation $\tilde{c}$ into the left sub network, the loss function is calculated by Equation (\ref{loss1})
\begin{equation}
\label{loss1}
loss_1(c,\tilde{c}) = \alpha * loss_{ngm}(c,\tilde{c}) + (1-\alpha) * loss_{ws}(c),
\end{equation}
\begin{equation}
loss_{ngm}(c,\tilde{c}) = max (0, 1-f^{ngm}(c)+f^{ngm}(\tilde{c})),
\end{equation}
\begin{equation}
loss_{ws}(c) = max (0,1-\phi(0)f^{ws}_0(c)+\phi(1)f^{ws}_1(c)),
\end{equation}
where $\alpha$ is linear interpolation weight and $\phi(\cdot)$ is an indicator of the sentiment polarity of the center word in $c$,
\begin{equation}
\phi(j) = 
\begin{cases}
1  & \text{if } y[j]==1, \\
-1 & \text{if } y[j]==0.
\end{cases} ,
\end{equation}
where $y$ is the gold label of a word, while we use 2-dimension vector to represent $y$, i.e. the negative polarity as [1,0] and the positive polarity as [0,1]. The sentiment polarity is from existing sentiment lexicon, in our experiments, we use the lexicon from \cite{HuLiu-2036}. Remarkably, if the center word is not sentiment word, we only optimize the n-gram score.

\begin{figure*}[!ht]
\centering
\includegraphics[width=0.8\textwidth]{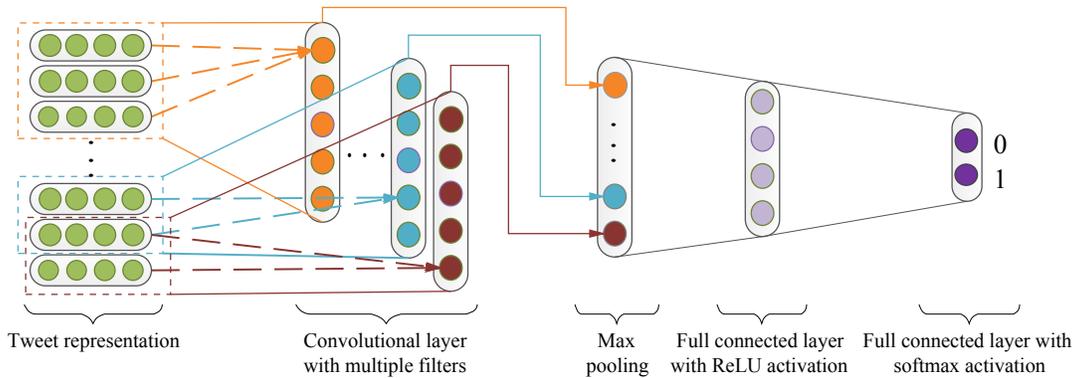}
\caption{Neural network architecture for sentiment classification}
\label{fig:classi}
\end{figure*}

\textbf{Twitter Level-Specific Layers}\hspace{5pt} For right sub network (modelling Twitter level sentiment), each linear transformation in shared unit is seen as a convolution operation on tweet text sequence. Subsequently, we use three pooling methods, \textit{max-pooling}, \textit{average-pooling} and \textit{min-pooling} on $e_1,e2,...,e_n$ to get fixed dimensional features $max(e)$, $avg(e)$ and $min(e)$. Then, we concatenate these features and feed them into a linear layer to get
\begin{equation}
a_2 = W^2_1 * [max(e) \oplus avg(e) \oplus min(e)] + b^2_2,
\end{equation}
where $W^2_1 \in R^{t*h \times h}$ and $b^2_1 \in R^h$ are the parameter of linear layer. Finally, the top softmax layer predicts the tweet sentiment
\begin{equation}
f^{ds} = softmax (a_2)
\end{equation}
The loss function of tweet level is 
\begin{equation}
\label{loss2}
loss_2(D) = - \sum_{k=\{0,1\}} g_k(D) {\log f^{ds}_k},
\end{equation}
where $g(\cdot)$ is the gold sentiment distribution of tweet on [\textit{positive}, \textit{negative}].

Since Equation (\ref{loss1}) and Equation (\ref{loss2}) optimize word level and tweet level information loss, respectively.
Our final optimization objective is to get an overall score, as follow
\begin{equation}
loss = \beta * loss_1(c,\tilde{c}) + (1-\beta) * loss_2(D)
\end{equation}
where $\beta$ is the trade-off coefficient between two levels.

\textbf{Model Training}\hspace{5pt} We train word embedding from both lexicon and massive distant-supervised tweets that is collected with positive (e.g. \#happy, \#joy, \#happyness) and negative (e.g. \#sadness, \#angry, \#frustrated) hashtag and emoticons (e.g. :( :-( : ( :) :-) : ) :D). We crawl tweets from March 1st, 2015 to April 31th, 2015. We tokenize each tweet with NLTK package, remove the @user, URLs, duplicates, quotes, spams and tweets written in language other than English. Finally, we collect 5M positive tweets and 5M negative tweets.

We use Stochastic Gradient Descent (SGD) to optimize the training target. For speeding up the training process, mini-batch training is commonly used in SGD. But, for word level training, there are a number of valid words\footnote{The word has enough left and right context words among the window settings.} in a tweet need to compute n-gram and sentiment loss. And there is one loss value in tweet level sentiment prediction. Moreover, for calculating tweet level sentiment prediction score, it must first compute the linear transformation $a_1$. Therefore, general mini-batch can not be used in our model. Here we use one trick which uses two batch sizes in training. The main batch size is for tweet level, and the second batch size is set as the window number in a tweet for word level training. In other words, the batch size is changeable for word level while batch size is fixed for tweet level through the training process. During training, we empirically set the window size as 3, the embedding dimension as 50, the length of hidden layer as 20, the main batch size as 32 and the learning rate as 0.01.

\subsection{Sentiment Classification with Sentiment-Enriched Word Embedding}
\label{sect:mswe}

After learning sentiment-specific word embedding, it can be used for sentiment classification by using existing supervised learning framework (e.g. SVM). Here, we use a neural network model to perform the classification task. The architecture of the model is showed in Figure \ref{fig:classi}. Firstly, a convolutional layer with multiple filters applied in the input tweet, which have been represented by using learned sentiment-specific word embedding. Then, a Max pooling layer takes the maximum value as the feature of each convolutional filter. The next hidden layer is a full connected layer with $ReLU$ activation, which is used for learning hidden feature representation. The final layer is a full connected layer with 2-dimension output that is used for predicting the positive/negative polarity distribution with $softmax$. We use dropout on input layer and hidden layer for regularization.  The classifier is trained by using back-propagation with AdaGrad to update parameters.

\section{Experimental Evaluation}
\subsection{Datasets and Settings}

\begin{table*}[!htb]
\center
\begin{tabular}{lrrrrrr}
\toprule 
\multirow{2}{*}{} &
\multicolumn{3}{c}{SemEval2013} & 
\multicolumn{3}{c}{CST}\\ 
\cmidrule(lr){2-4} \cmidrule(lr){5-7}
& Positive & Negative & Total& Positive & Negative & Total\\
\midrule
Train & 2,447 & 952 & 3,399 & 11,394 & 6,606 & 18,000 \\ 
Dev   & 575 & 340 & 915 & - & - & - \\ 
Test  & 1572 & 601 & 2,173 & CV & CV & CV \\
\bottomrule
\end{tabular}
\caption{Summary statistics for the datasets. CV means there was no standard train/test split and thus 10-fold CV was used.
\label{tab:ds}}
\end{table*}

\textbf{Datasets} \hspace{5pt} For demonstrating the effectiveness of the proposed method, we perform experiments on the following two datasets: 1) SemEval2013, which is a standard Twitter sentiment classification benchmark \cite{Nakov-2074}; 2) CST (Context-Sensitive Twitter), which is the latest benchmark for Twitter sentiment classification task \cite{RenZhang-2021}. \cite{RenZhang-2021} crawled basic opinion tweet and its context for evaluating their model, which utilized the contextual tweets as auxiliary data for Twitter sentiment classification. In our experiments, we only use the basic data rather than both basic and contextual tweets, because our model does not consider the contextual tweets at present. Table 1 provides detailed information about each dataset. Evaluation metric is Macro-F1 of positive and negative categories.

\textbf{Hyper-parameters} \hspace{5pt}  The hyper-parameter for specific task should carefully tune, for easily comparison of the experimental result, we use the unified setting that is chosen via a grid search on SemEval2013 developing dataset. There are seven hyper-parameters in the final model, including the network structure parameters (i.e. embedding dimension $D$, the length of hidden layer $H$, the convolutional filter size $S$ and filter number $N$) and the parameters for supervised training (i.e. the dropout rate of input layer $d_1$, the dropout rate of hidden $ReLU$ layer $d_2$ and the learning rate $\eta$ of AdaGrad). Table 2 reports all the hyper-parameters.

\begin{table}[!htb]
\center
\begin{tabular}{l|l}
\toprule 
Type & Value \\
\midrule
Network \\
structure &  $D$ = 50, $H$ = 200, $S$ = (2,3,4,5), $N$ = 30  \\ 
\midrule
Training  & $d_1$ = 0.8, $d_2$ = 0.7, $\eta$ = 0.01 \\
\bottomrule
\end{tabular}
\caption{Hyper-parameter values in the final model.
\label{tab:para}}
\end{table}

\subsection{Results of Comparison Experiments}
We compare our model with a number of state-of-the-art methods. The results of our models against other methods are listed in Table 3. All the methods fall into two categories: traditional classifier with various features and neural network classifier. In first category, SSWE achieves the best performance, which uses word embedding features that encode both n-gram and sentiment information. Because not explicitly exploiting sentiment information, C\&W and Word2vec features are relatively weak. NRC system performs better than other partners except SSWE, because of using sentiment lexicons and many manually designed features. For second category, neural network classifier can naturally use word embedding for classification. Both TSWE and CNNM-Local, which exploited other extra information rather than sentiment, achieved the current best performance in SemEval2013 and SST, respectively.  Under the same condition that only using sentiment information, our model performs better than them. Both our method and NRC utilize sentiment lexicons and achieve better performance, it demonstrates sentiment lexicon is still a strong resource for informal Twitter text sentiment classification.

\begin{table*}[!htb]
\center
\begin{tabular}{l|c|c}
\toprule 
Model & SemEval2013 & SST \\ 
\midrule
DistSuper + uni/bi/tri-gram \cite{GoBhayani-2071}	& 63.84 & - \\
SVM + uni/bi/tri-gram \cite{PangLee-1608}	 	& 75.06 & 77.42 \\
SVM + C\&W	\cite{TangWei-1500} 					& 75.89 & - \\
SVM + Word2vec \cite{TangWei-1500} 					& 76.31 & - \\
NBSVM	\cite{TangWei-1500} 					& 75.28 & - \\
RAE		\cite{TangWei-1500}						& 75.12 & - \\
NRC (Top System in SemEval)	\cite{Mohammad-2077}	& 84.73 & 80.24 \\
SSWE \cite{TangWei-1500}	 	 					& \textbf{84.98} & \textbf{80.68} \\
\midrule
TSWE \cite{RenZhang-2019}			& 85.34 & - \\
CNNM-Local \cite{RenZhang-2021}			& - & 80.90  \\
MSWE (Our model)						& \textbf{85.75} & \textbf{81.34} \\
\bottomrule
\end{tabular}
\caption{Results of our model against other methods. \textbf{DistSuper + uni/bi/tri-gram}: Lib-Linear Classifier with bag-of-ngram features trained on distant supervised tweets \cite{GoBhayani-2071}. \textbf{SVM + uni/bi/tri-gram}: SVM classifier with ngram features \cite{PangLee-1608}. \textbf{SVM + C\&W}: SVM classifier with C\&W word embedding features. \textbf{SVM + Word2vec}: SVM classifier with Word2vec word embedding features. NBSVM: a state-of-the-art performer which trades-off between Naive Bayes and NB-enhanced SVM \cite{WangManning-2076}. \textbf{RAE}: Recursive Autoencoders with pre-trained word vectors from Wikipedia \cite{SocherPennington-2075}. \textbf{NRC}: the top-performed system in SemEval 2013 Twitter sentiment classification track which incorporates diverse sentiment lexicons and many manually designed features \cite{Mohammad-2077}. \textbf{SSWE}: SVM classifier with SSWE word embedding features \cite{TangWei-1500}. \textbf{TSWE}: a neutral network classifier with topic and sentiment enriched word embedding features \cite{RenZhang-2019}. \textbf{CNNM-Local}: a neural network classifier which utilized the contextual tweets as auxiliary training resource \cite{RenZhang-2021}.
\label{tab:result}}
\end{table*}

\subsection{Effect of parameter $\beta$}

\begin{figure}[!htb]
\centering
\includegraphics[width=0.47\textwidth]{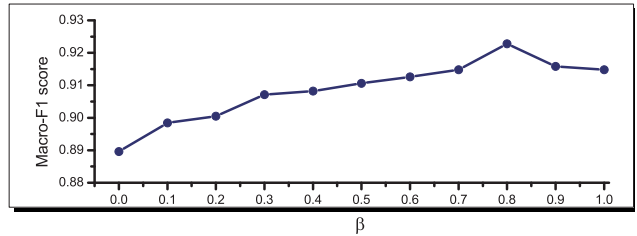}
\caption{Macro-F1 scores of MSWE with different $\beta$.}
\label{fig:para}
\end{figure}

As given in Equation (15), $\beta$ is the trade-off between word level and tweet level information. We tune $\beta$ on the development set of SemEval2013. For another coefficient $\alpha$, we follow \cite{TangWei-1500} to set it as 0.5. Figure \ref{fig:para} shows the macro-F1 scores of MSWE with different $\beta$ on SemEval2013 development set. It shows that MSWE gives better performance when $\beta$ = 0.8, which means giving more consideration to n-gram and word sentiment from lexicon. The model with $\beta$ = 1 stands for SSWE but the training sentiment label of word is from lexicon and $\beta$ = 0 stands for only using tweet level sentiment information. The model with $\beta$ = 0 gives lower performance, which shows the n-gram information is an indispensable evidence for Twitter sentiment classification. As a result, we set $\beta$ as 0.8 in our final experiments.

\section{Related Work}
For Twitter sentiment classification, many work follow traditional sentiment classification methods \cite{PangLee-1608}, which used machine learning methods to train a classifier for Twitter sentiment. Except common used text features, there are some distant supervised features can be utilized\cite{GoBhayani-2071}. Many studies use these massive noisy labelled tweets as training data or auxiliary source for Twitter sentiment classification \cite{HuTang-2073}. Unlike previous studies, our approach uses distant supervised information as well as lexicon knowledge for training word embedding, which is a combination of noisy-labelled resource and knowledge base.

There is a large body of work on word embedding learning \cite{MikolovSutskever-1335,CollobertWeston-1507,PenningtonSocher-1333}.
These models are based on word correlations within context windows. Recently, several methods integrate other information into word representation learning for specific tasks, which are more beneficial than common embedding \cite{LiuQiu-2055,RenZhang-2019,TangWei-1500,ZhouHe-2057}. 

For sentiment classification, \cite{TangWei-1500} proposed to integrate the sentiment information of tweets into neural network to learn sentiment specific word embedding. \cite{TangWei-1500} added a new optimization objective on the top layer of C\&W model \cite{CollobertWeston-1507}, and it is able to add more optimization objective, such as topic distribution \cite{RenZhang-2019}. However, it is unable to simultaneously integrate both word level and tweet level sentiment information into its optimization function. Therefore, we propose multi-level sentiment specific word embedding learning model to tackle this problem. The difference is 1) our method uses lexicon knowledge as supervised information in word level sentiment, while they use twitter overall sentiment as the label of all the words in that twitter; 2) we treat each context window in word level learning as a convolution operation, and then pool all the context windows into neural network to predict tweet level sentiment polarity, while they assigned the tweet overall sentiment label to each word instead of modelling twitter sentiment.

Our method uses similar schema as multi-task learning, which is firstly proposed in \cite{Caruana-2070}. For natural language processing community, a notable work about multi-task learning was proposed by \cite{CollobertWeston-2013}. In their model, each considered task shared the lookup tables as well as the first hidden layer, and the task was randomly selected for training in each round. \cite{LiuGao-1881} proposed to jointly train semantic classification and information retrieval, which have more shared layers between two tasks. Most of multi-task learning frameworks can be seen as a parameter sharing approach. However, these work aim to train the model for multi-tasks themselves, while our method aims to learn the shared embedding. Therefore, our method uses context window as the only convolution filter for Twitter level sentiment modelling. Our main target is to bring Twitter level sentiment information to word embedding but not to predict Twitter sentiment in MSWE. In our model, a shared unit not only represents a context composition window in word level but also a convolution window for tweet level.

\section{Conclusion}
This paper proposes to utilize both word sentiment label from lexicon and tweet sentiment label from distant supervised information for training sentiment-enriched word embedding. Because these two information are crossing word and tweet level, we develop a multi-level sentiment-enriched word embedding learning method. Our method naturally integrates word level information into tweet level as a convolution result, while simultaneously modelling word level n-gram and sentiment information. When using learned word embedding to Twitter sentiment classification, it achieves the best results in standard benchmarks. For our future work, we have plans to integrate more information that may enhance Twitter sentiment classification.

\bibliographystyle{aaai}
\bibliography{mtnnsa}

\end{document}